\documentclass{article}

\PassOptionsToPackage{numbers, compress}{natbib}
\usepackage[preprint]{neurips_2026}

\usepackage[utf8]{inputenc}
\usepackage[T1]{fontenc}
\usepackage{hyperref}
\usepackage{url}
\usepackage{booktabs}
\usepackage{amsfonts}
\usepackage{amsmath}
\usepackage{amssymb}
\usepackage{nicefrac}
\usepackage{microtype}
\usepackage[table]{xcolor}
\usepackage{graphicx}
\usepackage{subcaption}
\usepackage{multirow}
\usepackage{algorithm}
\usepackage{algpseudocode}
\usepackage{arydshln}
\usepackage{float}

\title{Nonlinear Bipolar Compensation: Handling Outliers in Post-Training Quantization}

\author{%
  Peilin Sun$^{1,2}$ \quad
  Jianxin Wu$^{1,2}$\thanks{J. Wu is the corresponding author.} \\
  $^{1}$State Key Laboratory of Novel Software Technology, Nanjing University, China \\
  $^{2}$School of Artificial Intelligence, Nanjing University, China \\
  \texttt{sunpl@lamda.nju.edu.cn, wujx2001@nju.edu.cn}
}

\begin{document}

\maketitle

\begin{abstract}
Network quantization has emerged as one of the most practical model compression techniques, which significantly reduces a model's memory and compute consumption by mapping floating-point numbers to low-bit representations. However, existing quantization methods typically suffer from the speed-accuracy tradeoff and limited generalization. To address these issues, recent compensation-based methods offer an efficient yet general solution by introducing additional lightweight linear layers into the quantized network. However, the accuracy of these methods suffers from their limited compensation capability and high sensitivity to outliers. In this paper, we propose Nonlinear Bipolar Compensation (NBC), a post-training quantization approach that introduces nonlinear compensation to reduce the effect of outliers. We further design Bipolar Logarithmic Transformation (BLT), which compresses outliers by mapping both the quantized input and the quantization error into a transformed space. A simple linear layer is then applied for compensation in the transformed space, preserving the efficiency of our method. Extensive experiments across various tasks, models, and quantization methods confirm the effectiveness, efficiency, robustness, and generality of our NBC approach.
\end{abstract}

\section{Introduction}
\label{sec:intro}
In recent years, with the development of deep learning, deep neural networks have demonstrated outstanding performance across diverse fields, including vision~\cite{dosovitskiy2020image,liu2021swin}, language~\cite{brown2020language, devlin2019bert}, and multimodal~\cite{radford2021learning} tasks, simultaneously raising an increasing demand for resources including computation, storage, memory, and bandwidth. Therefore, model compression~\cite{cheng2017survey} not only becomes a research focus in academia, but also plays a crucial role in real-world application scenarios with resource constraints and real-time inference requirements.

Network quantization~\cite{gholami2022survey} stands out among various model compression techniques for its high efficiency and hardware-friendliness. Based on whether retraining is required to recover accuracy after quantization, existing quantization methods can typically be divided into the following two categories: Quantization Aware Training (QAT)~\cite{liang2025gplq,li2022q, esser2019learned,chen2024efficientqat} and Post-Training Quantization (PTQ)~\cite{yuan2022ptq4vit,li2023repq,moon2024instance,zhong2025towards, nagel2020up, li2021brecq, wei2022qdrop,wu2025aphq}.

 QAT and optimization-based PTQ methods can typically achieve higher accuracy, but their high computational cost and long training process limit their practical deployment. In contrast, PTQ methods require only a small amount of unlabeled data for fast calibration but suffer a severe accuracy drop under low-bit quantization. Therefore, current quantization methods face a difficult tradeoff between accuracy and efficiency. At the same time, the generality of existing quantization approaches is limited, requiring the customization of quantization methods for different model architectures and tasks. To summarize, the field of network quantization still urgently needs an efficient, general, and practical method.

\begin{figure}
    \centering
    \begin{subfigure}[t]{0.40\textwidth}
        \centering
        \includegraphics[height=3.5cm,keepaspectratio]{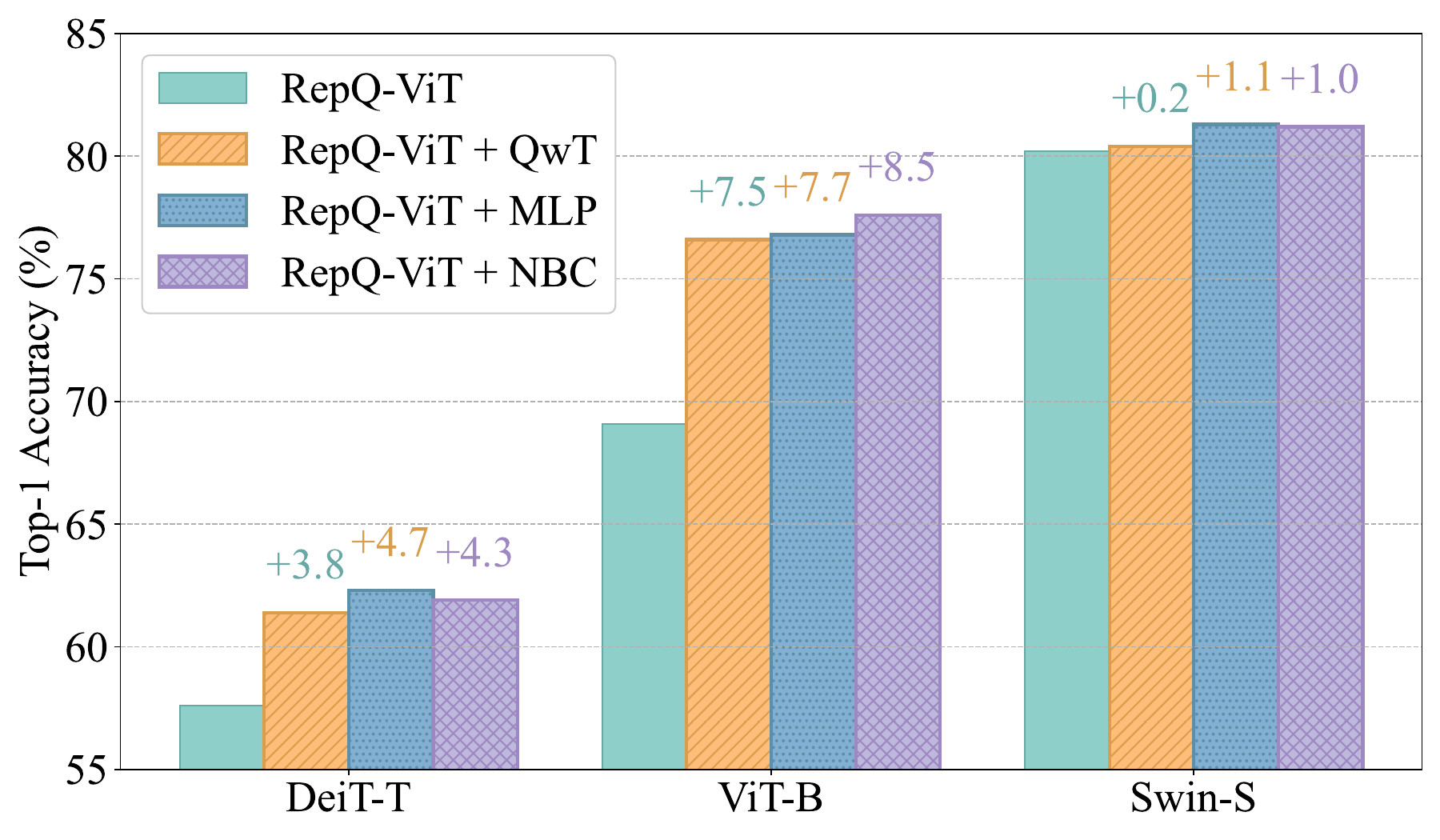}
        \caption{Nonlinearity improves accuracy}
        \label{fig:accuracy_plot}
    \end{subfigure}\hfill
    \begin{subfigure}[t]{0.58\textwidth}
        \centering
        \includegraphics[height=3.5cm,keepaspectratio]{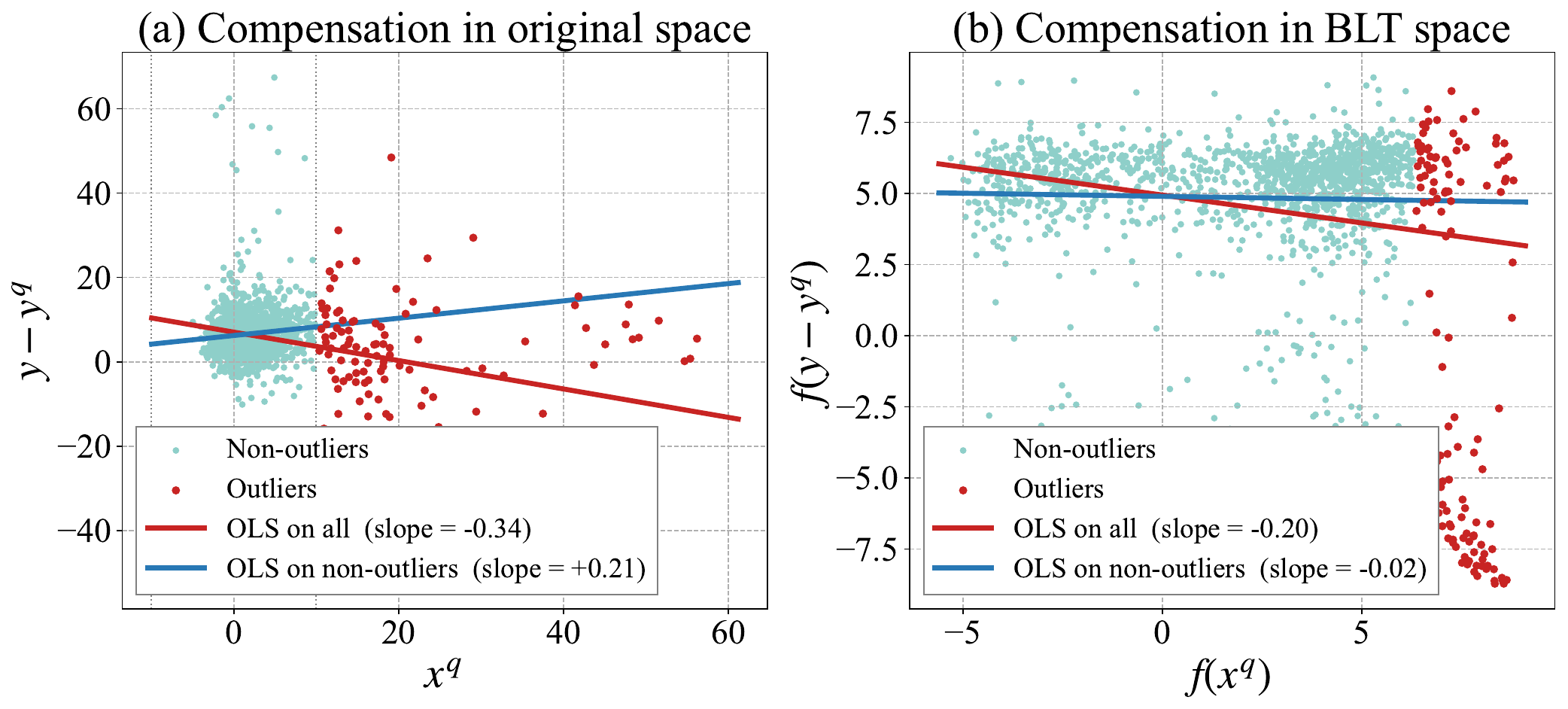}
        \caption{Outlier drag in OLS compensation}
        \label{fig:outlier_drag}
    \end{subfigure}
    \caption{The left figure (a) shows the ImageNet Top-1 accuracy of 4-bit quantized models, comparing nonlinear compensation and linear compensation. The MLP is a nonlinear compensation method that replaces the linear compensation module with a multilayer perceptron. Although it improves accuracy by introducing nonlinear compensation, it requires training and involves a large number of hyperparameters that must be adjusted. Please refer to the Supplementary Material for details on its implementation. The right figure (b) illustrates how outliers can pull the OLS-fitted linear compensation toward a small set of extreme activations, using a single $(x^q, y-y^q)$ channel pair in block~11 of W4A4 RepQ-ViT ViT-B. Outliers are defined as activations whose absolute value exceeds $10$. In the original space, OLS on all tokens is dragged opposite to OLS on inliers only, with a slope gap of $0.55$; after the BLT transform, the two lines become nearly parallel with a slope gap of $0.18$, motivating NBC's compensation in the BLT space.}
    \label{fig:accuracy}
\end{figure}

Recently, compensation-based methods~\cite{Fu_2025_CVPR,tang2025qwt} have introduced lightweight linear layers into quantized networks to mitigate quantization errors. While these approaches have achieved promising improvements in low-bit accuracy, their reliance on linear transformations introduces critical bottlenecks:

\begin{itemize}
    \item \textbf{Limited compensation capability.} These methods apply simple linear layers to compensate for the quantization loss. However, the information loss caused by quantization in deep neural networks is inherently highly nonlinear, making it difficult to be fully compensated by linear transformations. As a result, this has become the main reason that limits their accuracy. Figure~\ref{fig:accuracy_plot} illustrates the importance of nonlinearity in compensation by comparing nonlinear compensation and linear compensation.
    \item \textbf{High sensitivity to outliers.} In many modern networks (especially large language models and transformers), activations often contain outliers that significantly exceed the range of the majority of values~\cite{dettmers2022gpt3, darcet2023vision,xiao2023smoothquant}. In existing compensation-based methods, linear compensation modules are typically initialized by minimizing the residual sum of squares. Since this approach magnifies the effect of outliers, the resulting regression line tilts excessively toward them to minimize their large squared errors. This sacrifices optimal fit for the majority of the data, making these methods highly sensitive to outliers. Figure~\ref{fig:outlier_drag} illustrates how outliers can drag the Ordinary Least Squares (OLS) solution away from the majority of normal activations. We formally show that this failure mode is structural to linear OLS optimization in Appendix~\ref{app:ols_outlier}.
\end{itemize}

To address these issues, we propose a novel post-training quantization approach called Nonlinear Bipolar Compensation (NBC). Our NBC method introduces nonlinearity to better handle complex quantization loss and outliers, while keeping simplicity and efficiency.

NBC introduces a nonlinear function $f(x)$ and its inverse $f^{-1}(x)$. $f(x)$ maps the quantized input $x^q$ and the quantization error $y-y^q$ into a new space. A linear layer is then utilized to compensate for the quantization error in this new space. Finally, the result is mapped back to the original space via $f^{-1}(x)$ to achieve error compensation.

To compress outliers, we propose the Bipolar Logarithmic Transformation (BLT), which serves as the specific implementation of the nonlinear function $f(x)$. BLT employs a linear mapping near the origin and applies logarithmic compression to values farther from zero. BLT addresses the limitation of the standard log transformation for data containing negative values, while preserving its benefit of mitigating outliers.

Our NBC method offers four key advantages:
\begin{itemize}
    \item \textbf{Effectiveness.} NBC delivers substantial accuracy gains in the low-bit regime, e.g., $+4.6\%$ average top-1 over the RepQ-ViT~\cite{li2023repq} baseline at W4A4 ImageNet classification, and consistently outperforms the linear compensation method QwT~\cite{Fu_2025_CVPR}.
    \item \textbf{Efficiency.} The compensation parameters are obtained via a closed-form solution, requiring no backpropagation. Calibration completes within a few minutes. Our NBC module adds only roughly 4\% overhead relative to the full-precision backbone.
    \item \textbf{Robustness.} By transforming both the quantized input and the quantization error through BLT, NBC reduces compensation error on both outlier and non-outlier activations.
    \item \textbf{Generality.} NBC consistently improves accuracy across diverse tasks (vision, language, and multimodal recognition) and PTQ methods (RepQ-ViT~\cite{li2023repq}, AdaLog~\cite{wu2024adalog}, Q-DiT~\cite{chen2025q}, DuQuant~\cite{lin2024duquant}).
\end{itemize}

We summarize the main contributions of this paper as follows.
\begin{itemize}
    \item We analyze existing compensation-based PTQ methods and identify two key limitations: their limited capability to compensate for
  nonlinear quantization loss and their high sensitivity to \emph{outliers in activations}.
    \item To overcome the limitations of existing compensation-based methods, we propose the Nonlinear Bipolar Compensation (NBC) method. NBC
    introduces nonlinearity while maintaining high efficiency and simplicity. Notably, all parameters in the compensation module can still be
    obtained via a closed-form solution.
    \item We propose Bipolar Logarithmic Transformation (BLT), which overcomes the standard logarithmic transformation's limitation (\emph{unable to handle negative values}), thereby efficiently compressing outliers in activations.
\end{itemize}
To evaluate our method, we have conducted extensive experiments across tasks in various domains, including vision, language, and multi-modality. The experimental results demonstrate the effectiveness, efficiency, robustness, and generality of our approach.

\section{Related Works}
\label{sec:related}

Network quantization~\cite{gholami2022survey} compresses neural networks by representing the original floating-point weights and activations with low-bit values. Existing quantization methods can be classified into Quantization-Aware Training (QAT), which requires retraining, and Post-Training Quantization (PTQ), which does not. Although QAT methods~\cite{liang2025gplq,li2022q,esser2019learned,chen2024efficientqat} achieve high accuracy, they require labeled data and incur substantial training overhead.

In contrast, PTQ methods~\cite{yuan2022ptq4vit,li2023repq, moon2024instance,zhong2025towards, nagel2020up, li2021brecq, wei2022qdrop, wu2025aphq,jiang2026uq, yang2024dopq, wu2025fima} are considered an efficient and practical solution because they only require a small amount of unlabeled data to calibrate the quantization parameters and eliminate the need for retraining. They can be further divided into two categories: calibration-only methods~\cite{yuan2022ptq4vit,li2023repq,moon2024instance,zhong2025towards} and optimization-based methods~\cite{nagel2020up, li2021brecq, wei2022qdrop,yang2024dopq, wu2025aphq, wu2025fima}.

Calibration-only methods adjust quantization parameters only during the calibration stage, leading to higher efficiency and lower deployment cost. RepQ-ViT~\cite{li2023repq} decouples quantization and inference, employing scale reparameterization to convert high-accuracy quantizers used during quantization into hardware-friendly quantizers for inference. ERQ~\cite{zhong2025towards} models the minimization of the activation quantization error as a ridge regression problem and solves it by updating the full-precision weights. AdaLog~\cite{wu2024adalog} proposes an adaptive logarithm quantizer and a fast progressive combining search strategy. UQ-ViT~\cite{jiang2026uq} mitigates quantization errors from extreme ViT activations using hardware-friendly uniform quantization. However, these methods still suffer from significant accuracy degradation under low-bit quantization, and their reliance on non-uniform quantizers is hardware-unfriendly, further limiting practical deployment.

Optimization-based methods~\cite{nagel2020up, li2021brecq, wei2022qdrop, yang2024dopq, wu2025aphq, wu2025fima} introduce a block-wise reconstruction stage beyond calibration to recover accuracy. DopQ-ViT~\cite{yang2024dopq} addresses post-Softmax distribution mismatch and post-LayerNorm scaling-factor outliers using TanQ and MOSF, respectively. APHQ-ViT~\cite{wu2025aphq} proposes an Average Perturbation Hessian loss for reconstruction and an MLP reconstruction strategy to alleviate post-GELU activation quantization errors. FIMA-Q~\cite{wu2025fima} reveals the linear relationship between KL divergence and the Fisher Information Matrix, and proposes an improved FIM approximation for more accurate reconstruction loss. Although these methods achieve higher accuracy, they rely on numerous hyperparameters and substantial training overhead, which limits their practical deployment.

\section{Method}
\subsection{Preliminaries}

\textbf{Uniform Quantization.} Among various quantization methods, uniform quantization is widely employed owing to its simplicity and hardware friendliness. For a given bit width $b$, uniform quantization quantizes a floating-point number $x$ into the low-bit representation $x^{\text{q}}$. This process can be formalized as follows:
\begin{equation}
x^{\text{q}} = \text{clip} \left( \left\lfloor \frac{x}{s} \right\rceil + z, 0, 2^b - 1 \right)\,.
\end{equation}
Here, $\lfloor \cdot \rceil$ denotes the rounding operation, $\text{clip}(\cdot, a, b)$ restricts the input value to the range $[a, b]$, $s \in \mathbb{R}^+$ represents the quantization scale, and $z \in \mathbb{Z}$ denotes the zero-point offset. The quantization scale and zero-point offset are determined by the following equations:
\begin{align}
s &= \frac{\max(x) - \min(x)}{2^b - 1} \label{eq:s_def} \,,\\
z &= \text{clip} \left( \left\lfloor -\frac{\min(x)}{s} \right\rceil, 0, 2^b - 1 \right) \,. \label{eq:z_def}
\end{align}

The low-precision representation can be converted back to floating-point numbers via dequantization:
\begin{equation}
\hat{x} \approx s \times (x^{\text{q}} - z).
\end{equation}

\textbf{Blockwise Linear Compensation in QwT.} Observing the information loss introduced by quantization, QwT~\cite{Fu_2025_CVPR} suggests employing a lightweight linear layer to correct this information loss blockwise, demonstrating significant effectiveness in various tasks.

Modern neural networks are typically composed of blocks. For a given block, $x^q$ is the quantized input. The output before and after quantization is denoted by $y$ and $y^q$, respectively. QwT employs a simple linear layer to correct the quantization loss, which can be formalized as follows:
\begin{equation}
y^{\text{qwt}} = y^q + Wx^q+b.
\label{eq:qwt}
\end{equation}
The weight matrix $W \in \mathbb{R}^{d_{out} \times d_{in}}$ and bias vector $b \in \mathbb{R}^{d_{out}}$ can be found in closed-form.

Although QwT significantly improves the accuracy of existing PTQ methods, the highly nonlinear quantization loss is difficult to fully compensate using such a simple linear layer.

\subsection{Key Ideas in Our NBC}
\label{sec:key_ideas}
To address this issue, we attempt to \emph{introduce nonlinearity to better compensate} for complex quantization loss while retaining the advantages of its \emph{closed-form solution, plus both quantization and inference efficiency}. We introduce a \emph{nonlinear} transformation function $f(x)$ that maps the quantized input $x^q$ to a new space. Subsequently, in this transformed space, we perform compensation via a simple linear layer. Finally, the compensation is mapped back to the original output space through the inverse function $f^{-1}(x)$, thus achieving a nonlinear correction of the quantization error. It can be formalized as follows:
\begin{equation}
y^{\text{nbc}} = y^q + f^{-1}(Wf(x^q)+b).
\label{nbc}
\end{equation}
Here, $f^{-1}(x)$ is the inverse function of $f(x)$, while $W \in \mathbb{R}^{d_{out} \times d_{in}}$ and $b\in \mathbb{R}^{d_{out}}$ are the weight matrix and bias vector, respectively. It is worth noting that a closed-form solution can still be obtained through linear regression on $f(x^q)$ and $f(y-y^q)$, providing the advantages of simplicity and efficiency.

Our key insight is that \textit{the outliers in $x^q$ and $y-y^q$ pose severe challenges to linear compensation}. Therefore, our objective is to design a proper function $f(x)$ that can reduce the impact of outliers through a nonlinear transformation.

Logarithmic transformations~\cite{osborne2002notes} are widely employed to deal with outliers. However, as $x^q$ and $y-y^q$ may contain negative values, the standard logarithmic transformation is not applicable for our purposes.

\begin{figure}[tbp]
    \centering
    \begin{subfigure}[t]{0.35\textwidth}
        \centering
        \includegraphics[width=\linewidth]{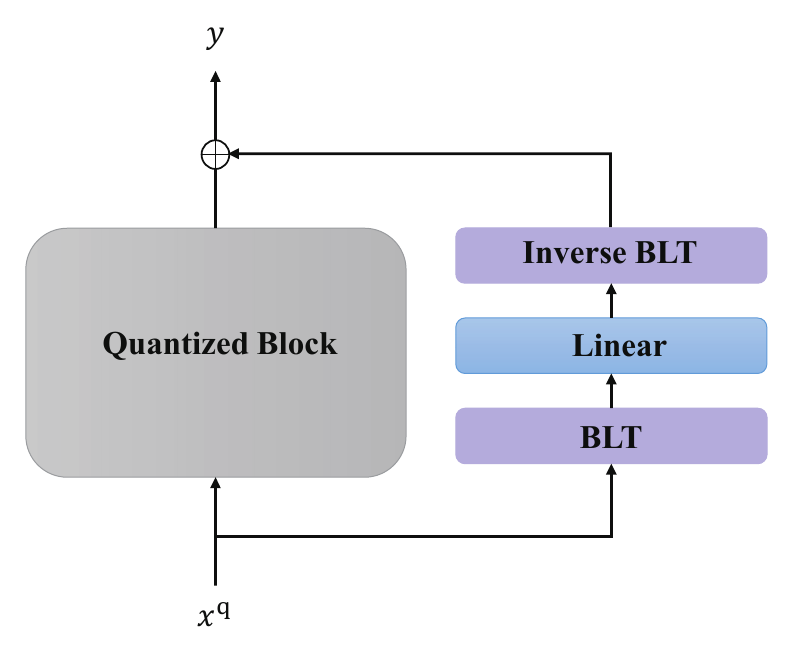} 
        \caption{NBC illustration}
        \label{fig:nbc}
    \end{subfigure}%
    \hfill 
    \begin{subfigure}[t]{0.60\textwidth} 
        \centering
        \includegraphics[width=\linewidth]{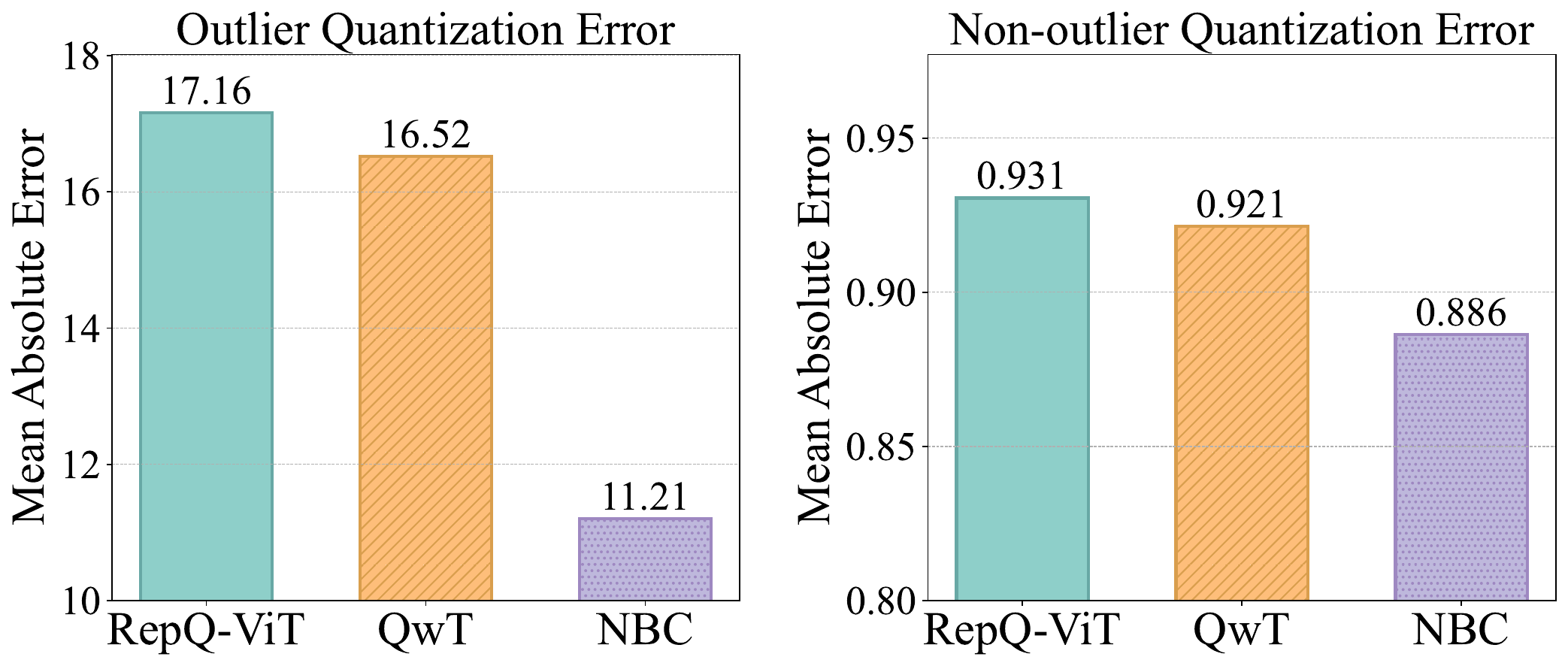}
        \caption{Quantization error comparison}
        \label{fig:compensation}
    \end{subfigure}
    
    \caption{The left figure (a) shows an NBC illustration on a single block. NBC utilizes BLT to nonlinearly transform quantized input and quantization error. A lightweight linear layer is then employed in this new space to compensate for quantization error. The right figure (b) shows a comparison of quantization error of RepQ-ViT~\cite{li2023repq}, QwT~\cite{Fu_2025_CVPR}, and our NBC method on outliers (activations with absolute value greater than $10$) and non-outliers for activations in the 11th block of ViT-B~\cite{dosovitskiy2020image} under 4-bit quantization. We use the mean absolute error between $y$ and $y^q$ to measure the quantization error.}
    \label{fig:main_figure}
\end{figure}

To address this issue, we propose the Bipolar Logarithmic Transformation (BLT). We employ it as the implementation of the nonlinear function $f(x)$ in Equation~\ref{nbc} and call the resulting nonlinear compensation method Nonlinear Bipolar Compensation (NBC). The NBC structure is illustrated in Figure~\ref{fig:nbc}.

Figure~\ref{fig:compensation} compares the quantization errors of our NBC method and existing compensation-based methods across the domains of non-outliers and outliers, demonstrating that NBC achieves better compensation accuracy on both outliers and non-outliers. It is worth noting that NBC achieves a more significant effect on outliers than on non-outliers, demonstrating its superior performance in handling outliers.

\subsection{Bipolar Logarithmic Transformation}
\label{sec:BLT}

\begin{figure}[tbp]
    \centering
    \includegraphics[width=0.48\textwidth]{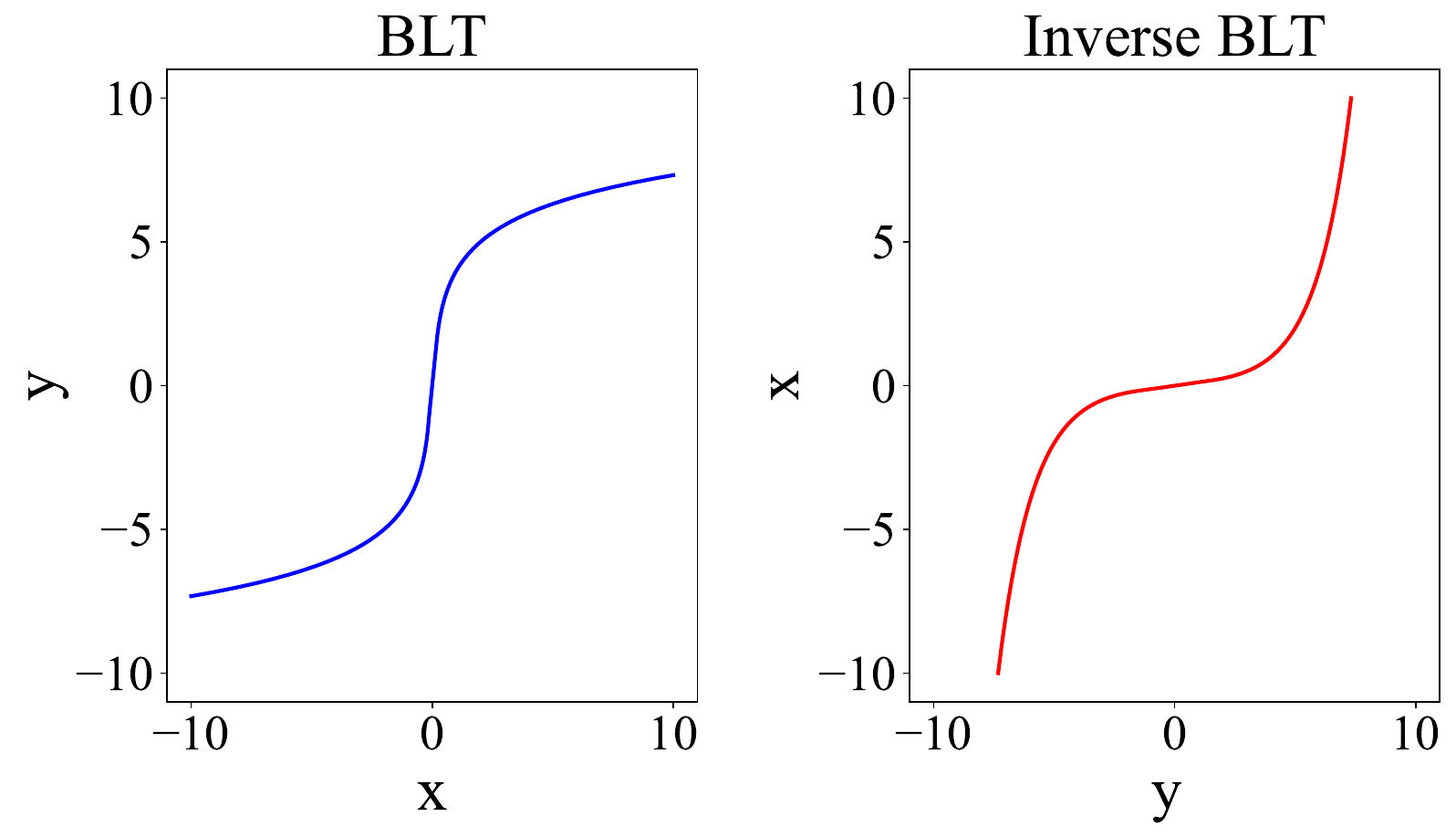}
    \caption{Plots for BLT and Inverse BLT.}
    \label{fig:BLT} 
\end{figure}

Our Bipolar Logarithmic Transformation (BLT) partitions the input domain into three distinct regions based on a magnitude threshold: values with small magnitudes undergo a linear transformation, while larger positive and negative values utilize a logarithmic transformation. This strategy successfully extends the benefits of logarithmic transformation to \emph{input data with negative values}, which can be formalized as:
\begin{equation}
f(x) =
\begin{cases}
\log_2(x)+N+1, & x > 2^{-N} \\
2 ^ N \times x, & 2^{-N} \geq x \geq -2^{-N}\\
-\log_2(-x)- N -1, & x < -2^{-N}
\end{cases}
\,.
\end{equation}
Here, $N$ denotes the hyperparameter used to adjust the threshold. Plots for the BLT and Inverse Bipolar Logarithmic Transformation (Inverse BLT) functions are shown in Figure~\ref{fig:BLT}. For values in the range $[-2^{-N}, 2^{-N}]$, $f(x)$ scales them linearly and makes sure $f(x)$ resides in $[-1, 1]$. For positive values $x>2^{-N}$, $f(x)$ transforms them using a logarithmic transformation plus an \emph{offset} $N+1$, such that $f(x)$ resides in the non-overlapping range $(1,+\infty)$. More importantly, for negative values $x<-2^{-N}$, $f(x)$ maps them to the range $(-\infty,-1)$ with the help of a negative offset $-N-1$. Therefore, our $f(x)$ can handle both positive and negative values logarithmically, hence the name \emph{bipolar} logarithmic transformation (BLT).

The hyperparameter $N$ determines the interval boundaries for input values. For example, $N=5$ yields a central linear region of $[-0.031, 0.031]$. We use a single global $N$ shared across all blocks of the model. To select $N$, we use a Feature Loss based Local Search (FLS) algorithm (see Appendix~\ref{app:fls} for the full algorithm) starting from an empirical $N_{\text{init}}$ and exploring its discrete neighborhood with step size $s$. The search converges within a few steps in practice.

\section{Experiments}

We first evaluate our NBC method on image classification and image generation. Subsequently, we extend our experiments to multimodal recognition and large language models. Object detection and instance segmentation results are reported in Appendix~\ref{app:detection} due to space constraints. Please refer to the Supplementary Material for more implementation details and additional ablation studies.

\subsection{Experiments on Image Classification}

\textbf{Settings.} For classification, we evaluated our method on the ImageNet~\cite{deng2009imagenet} dataset, employing diverse backbone network structures including ViT~\cite{dosovitskiy2020image}, DeiT~\cite{touvron2021training}, and Swin~\cite{liu2021swin}. Since NBC targets the fast calibration regime, we compare it against calibration-only PTQ baselines, including RepQ-ViT~\cite{li2023repq}, and the compensation-based competitor QwT~\cite{Fu_2025_CVPR}, which can also be integrated with various PTQ methods. We randomly sampled 512 images from the training set for NBC parameter initialization and FLS hyperparameter search. In all experiments, the parameters in the NBC module are stored using FP16 precision. The rest of the experimental setup was consistent with prior work~\cite{Fu_2025_CVPR}.

\begin{table}[tbp]
\centering
\caption{Quantization results on ImageNet~\cite{deng2009imagenet} classification. `Size' (MB) is the storage size. `Top-1' (\%) is the top-1 accuracy. `*' denotes that the NBC parameters are stored in INT8 instead of FP16.}
\label{tab:cls_main}
\begin{tabular}{l c l c c c c}
\toprule
\multirow{2}{*}{Network} & \multirow{2}{*}{Full Prec.} & \multirow{2}{*}{Method} & \multicolumn{2}{c}{W4/A4} & \multicolumn{2}{c}{W6/A6} \\
\cmidrule(lr){4-5} \cmidrule(lr){6-7}
& & & Size & Top-1 & Size & Top-1 \\
\midrule
\multirow{4}{*}{DeiT-T} & \multirow{4}{*}{72.1} & RepQ-ViT~\cite{li2023repq} & 3.3 & 57.6 & 4.6 & 70.9 \\
& & RepQ-ViT + QwT & 4.2 & 61.4 & 5.5 & 71.2 \\
& & RepQ-ViT + NBC & 4.2 & \textbf{61.9} & 5.5 & 71.2 \\
& & RepQ-ViT + NBC* & 3.7 & 61.8 & 5.1 & \textbf{71.3} \\
\midrule
\multirow{4}{*}{ViT-B} & \multirow{4}{*}{84.5} & RepQ-ViT~\cite{li2023repq} & 44.9 & 69.1 & 66.2 & 83.8 \\
& & RepQ-ViT + QwT & 59.1 & 76.6 & 80.4 & 83.8 \\
& & RepQ-ViT + NBC & 59.1 & \textbf{77.6} & 80.4 & \textbf{83.9} \\
& & RepQ-ViT + NBC* & 52.0 & \textbf{77.6} & 73.3 & \textbf{83.9} \\
\midrule
\multirow{4}{*}{Swin-S} & \multirow{4}{*}{83.2} & RepQ-ViT~\cite{li2023repq} & 25.8 & 80.2 & 38.0 & 82.8 \\
& & RepQ-ViT + QwT & 33.7 & 80.4 & 45.9 & \textbf{82.9} \\
& & RepQ-ViT + NBC & 33.7 & \textbf{81.2} & 45.9 & \textbf{82.9} \\
& & RepQ-ViT + NBC* & 29.7 & 81.0 & 42.0 & \textbf{82.9} \\
\bottomrule
\end{tabular}
\end{table}

\textbf{Main Results.} Table~\ref{tab:cls_main} shows that NBC consistently improves accuracy across various backbone architectures and outperforms QwT. Specifically, NBC improves accuracy by 2.4\% on average compared to the baseline and even achieves an average gain of 4.6\% under 4-bit quantization, indicating that our method is especially effective in low-bit quantization scenarios. We further quantize the NBC parameters from FP16 to INT8 using a simple per-channel min-max algorithm, and the additional storage of the compensation module is roughly halved while the top-1 accuracy stays within $0.2\%$ of the FP16 NBC counterpart, showing that NBC itself can be safely 8-bit quantized to reduce storage with negligible accuracy loss.

\begin{table}[tbp]
    \centering
    \caption{Quantization results across various PTQ methods on ImageNet~\cite{deng2009imagenet} with the ViT-B~\cite{dosovitskiy2020image} backbone under 4-bit quantization.}
    \label{tab:cls_ptq}
    \begin{tabular}{ll c c c}
        \toprule
        Network & Method & \#Bits & Size & Top-1 \\
        \midrule
        \multirow{7}{*}{ViT-B} & Full-precision & 32/32 & 346.3 & 84.5 \\
        \cdashline{2-5}
        & RepQ-ViT~\cite{li2023repq} & 4/4 & 44.9 & 69.1 \\
        & RepQ-ViT + QwT & 4/4 & 59.1 & 76.6 \\
        & RepQ-ViT + NBC & 4/4 & 59.1 & \textbf{77.6} \\
        \cdashline{2-5}
        & AdaLog~\cite{wu2024adalog} & 4/4 & 44.9 & 79.7 \\
        & AdaLog + QwT  & 4/4 & 59.1 & 80.1 \\
        & AdaLog + NBC & 4/4 & 59.1 & \textbf{80.3} \\
        \bottomrule
    \end{tabular}
\end{table}

\textbf{Results on different PTQ methods.} We further integrated NBC with various PTQ approaches to verify its generality. Table~\ref{tab:cls_ptq} presents the experimental results of integrating NBC with RepQ-ViT~\cite{li2023repq} and AdaLog~\cite{wu2024adalog} on the ViT-B~\cite{dosovitskiy2020image} backbone under 4-bit quantization. The results demonstrate that NBC consistently improves accuracy across various PTQ methods and surpasses QwT's performance.

\subsection{Experiments on Image Generation}

\textbf{Settings.} For the image generation task, we employed the DiT-XL/2 model under 256x256 resolution and used Q-DiT~\cite{chen2025q} as the baseline method. We compare our method with RepQ-ViT~\cite{li2023repq}, GPTQ~\cite{frantar2022gptq} and Q-DiT~\cite{chen2025q} specifically designed for diffusion models. To achieve a balance between speed and accuracy, we used the DDIM sampler with 50 steps and applied a classifier-free guidance (cfg) value of 1.5. Following QwT~\cite{Fu_2025_CVPR}, we make the assumption that the quantization error primarily depends on the input token $x$ and is minimally affected by the time step $t$. Therefore, we set $t=0$ for the initialization of the NBC module.

\textbf{Main Results.} Table~\ref{tab:gen_main} presents the experimental results of our NBC method on image generation tasks. In both the W4A8 and W8A8 settings, NBC significantly improves the baseline PTQ method, resulting in average improvements of 0.33 for FID and 5.4 for IS.
Compared to the baseline PTQ method, NBC requires only approximately 4\% of additional model size.

\subsection{Experiments on Multimodal Recognition}

\textbf{Settings.} For multimodal recognition tasks, we conducted experiments on the CLIP~\cite{radford2021learning} model. In our experiments, we selected ViT-B/32~\cite{dosovitskiy2020image} as the vision encoder and a 12-layer Transformer~\cite{vaswani2017attention} as the text encoder. We adopted the quantization method based on RepQ-ViT~\cite{li2023repq} developed in QwT~\cite{Fu_2025_CVPR} as the baseline. We adopted the zero-shot classification top-1 accuracy on ImageNet~\cite{deng2009imagenet} as the evaluation metric. We conducted our experiments using two quantization settings: quantizing only the image encoder and quantizing both the image and text encoders.

\textbf{Main Results.} The results of our NBC method on multimodal recognition tasks are shown in Table~\ref{tab:mm_main}. The experimental results demonstrate that integrating NBC significantly improves the accuracy of the quantized CLIP model across various quantization scenarios. Specifically, when only the image encoder is quantized, NBC yields an average accuracy improvement of 0.9\% while introducing an overhead of only about 4\% in additional model size. When both the image and text encoders are simultaneously quantized, the accuracy gain from NBC reaches 16.4\%, with approximately 9\% additional model size overhead.

\begin{table}[tbp]
  \centering
  \begin{minipage}[t]{0.48\linewidth}
    \centering
    \caption{Quantization results of DiT-XL/2~\cite{peebles2023scalable}. We adopt the Fr\'{e}chet Inception Distance (FID)~\cite{heusel2017gans} and Inception Score (IS) as our evaluation metrics. $\downarrow$ ($\uparrow$) indicates that lower (higher) values are better.}
    \label{tab:gen_main}
    \setlength{\tabcolsep}{4pt}
    \resizebox{\linewidth}{!}{
      \begin{tabular}{l c c c c}
      \toprule
      Method & \#Bits & Size (MB) & FID ($\downarrow$) & IS ($\uparrow$) \\
      \midrule
      Full-precision  & 16/16 & 1349 & \phantom{31}5.32 & 236.17 \\
      \midrule
      RepQ-ViT~\cite{li2023repq}        & 8/8   & \phantom{1}677  & \phantom{31}5.46 & 234.74 \\
      GPTQ~\cite{frantar2022gptq}       & 8/8   & \phantom{1}690  & \phantom{31}5.90 & 218.90 \\
      Q-DiT~\cite{chen2025q}            & 8/8   & \phantom{1}683  & \phantom{31}5.45 & 236.91 \\
      Q-DiT + NBC     & 8/8   & \phantom{1}707  & \phantom{31}\textbf{5.41} & \textbf{240.40} \\
      \midrule
      RepQ-ViT~\cite{li2023repq}        & 4/8   & \phantom{1}339  & 319.68 & \phantom{31}2.20 \\
      GPTQ~\cite{frantar2022gptq}       & 4/8   & \phantom{1}351  & \phantom{31}9.04   & 166.35 \\
      Q-DiT~\cite{chen2025q}            & 4/8   & \phantom{1}347  & \phantom{31}6.75   & 208.38 \\
      Q-DiT + NBC     & 4/8   & \phantom{1}361  & \phantom{31}\textbf{6.12}   & \textbf{215.70} \\
      \bottomrule
      \end{tabular}
    }
  \end{minipage}\hfill 
  \begin{minipage}[t]{0.48\linewidth}
    \centering
    \caption{Zero-shot image classification results on ImageNet~\cite{deng2009imagenet}. `Setup' indicates the two quantization settings used in our experiments: quantizing only the image encoder and quantizing both the image and text encoders simultaneously.}
    \label{tab:mm_main}
    \setlength{\tabcolsep}{2pt}
    \resizebox{\linewidth}{!}{
      \begin{tabular}{l l c c c}
      \toprule
      Setup & Method & \#Bits & Size & Top-1 \\
      \midrule
      \multirow{4}{*}{Vision} 
      & Full-precision      & 32/32 & 607.2 & 63.4 \\
      \cdashline{2-5}
      & RepQ-ViT~\cite{li2023repq}            & 6/6   & 323.5 & 59.2 \\
      & RepQ-ViT + NBC        & 6/6   & 336.8 & \textbf{60.7} \\
      \cdashline{2-5}
      & RepQ-ViT~\cite{li2023repq}           & 8/8   & 345.3 & 62.9 \\
      & RepQ-ViT + NBC        & 8/8   & 359.5 & \textbf{63.1} \\
      \midrule
      \multirow{5}{*}{\shortstack{Vision \\ \& Text}} 
      & Full-precision      & 32/32 & 607.2 & 63.4 \\
      \cdashline{2-5}
      & RepQ-ViT~\cite{li2023repq}          & 6/6   & 200.8 & 29.8 \\
      & RepQ-ViT + NBC        & 6/6   & 221.3 & \textbf{44.3} \\
      \cdashline{2-5}
      & RepQ-ViT~\cite{li2023repq}           & 8/8   & 232.1 & 38.7 \\
      & RepQ-ViT + NBC        & 8/8   & 252.6 & \textbf{57.0}\\
      \bottomrule
      \end{tabular}
    }
  \end{minipage}

\end{table}

\subsection{Experiments on Large Language Models}

\textbf{Settings.} For large language models, we employed LLaMA2-7B~\cite{touvron2023llama} and selected DuQuant~\cite{lin2024duquant} as the baseline PTQ method for W4A4 quantization. Our method is also compatible with other quantization approaches such as GPTQ~\cite{frantar2022gptq} and AWQ~\cite{lin2024awq}. 

Following the settings of DuQuant~\cite{lin2024duquant}, we evaluated perplexity on the WikiText2~\cite{merity2016pointer} and C4~\cite{raffel2020exploring} datasets for the language generation task. For commonsense QA tasks, we assessed zero-shot accuracy using the PIQA~\cite{bisk2020piqa}, ARC~\cite{clark2018think}, BoolQ~\cite{clark2019boolq}, HellaSwag~\cite{zellers2019hellaswag}, and WinoGrande~\cite{sakaguchi2021winogrande} datasets.

\textbf{Main Results.} Table~\ref{tab:llm_main} reports NBC results on quantized LLMs, including perplexity on WikiText2 and C4 and average accuracy on five commonsense QA datasets. NBC consistently improves the baseline across all metrics. Notably, DuQuant~\cite{lin2024duquant} is a rotation-based PTQ method that suppresses outliers via block-diagonal orthogonal transforms; the fact that NBC still yields consistent gains on top of it shows that nonlinear compensation is complementary to rotation-based outlier handling and the two can be stacked.

\begin{table}[tbp]
\centering
\caption{Quantization results of LLaMA2-7B~\cite{touvron2023llama}. `Size' (GB) indicates the model's storage size. `W2' is the abbreviation for WikiText2. $\downarrow$ ($\uparrow$) indicates that lower (higher) values are better.}
\label{tab:llm_main}
\setlength{\tabcolsep}{1.5pt}
\begin{tabular}{l c c c c c c}
\toprule
Method & \#Bits & Size  & W2 ($\downarrow$) & C4 ($\downarrow$) & QA Avg ($\uparrow$)\\
\midrule
Full-precision     & 16/16 & 13.54  & 5.47 & 6.97 & 63.72 \\
\midrule
DuQuant            & 4/4   & 3.89  & 6.28 & 7.90 & 60.57 \\
DuQuant + NBC      & 4/4   & 4.97  & \textbf{6.03} & \textbf{7.69} & \textbf{61.47} \\
\bottomrule
\end{tabular}
\end{table}

\subsection{Ablation Studies}
\textbf{Inference Efficiency of NBC.} To evaluate the practical deployment efficiency of our proposed method,
we measured the inference latency and model size of various architectures under the W8A8 setting on a GPU, as summarized in Table~\ref{tab:latency_comparison}. We choose W8A8 over the W4A4 setting used in our accuracy experiments because, to the best of our knowledge, no production-grade inference framework (TensorRT~\cite{TensorRT}, ONNX Runtime~\cite{ONNXRuntime}, TVM~\cite{chen2018tvm}, FasterTransformer~\cite{FasterTransformer}) currently provides W4A4 GEMM kernels for Vision Transformers; existing W4A4 kernels (e.g., Marlin~\cite{frantar2025marlin}, Atom~\cite{zhao2024atom}) target only LLMs.

The experiments employed the tensor-wise Percentile~\cite{li2019fully} method as the baseline, which is the standard W8A8 PTQ baseline supported by TensorRT~\cite{TensorRT}. Compared with the full-precision models, NBC achieves a favorable accuracy-efficiency trade-off, reducing inference latency by $73\%$ and model size by $71\%$ on average while incurring only a $0.6\%$ Top-1 accuracy drop. In contrast, the Percentile baseline alone, although marginally cheaper, suffers a $3.6\%$ accuracy drop on average, primarily driven by an $8.9\%$ loss on ViT-B. NBC's additional $4\%$ resource cost therefore recovers nearly all of the quantization-induced accuracy loss, delivering near-full-precision quality at roughly one-third the FP latency and footprint and demonstrating the strong practicality of NBC for W8A8 deployment.

\begin{table}[tbp]
\centering
\caption{8-bit quantization results. Latency (ms) is measured on a single RTX 3090 GPU with a batch size of 64, utilizing Nvidia's TensorRT~\cite{TensorRT} toolkit for deployment.}
\label{tab:latency_comparison}
\begin{tabular}{l l c c c}
\toprule
Network & Method & Size & Latency & Top-1 \\
\midrule
\multirow{3}{*}{DeiT-T} 
 & Full-precision & 22.9 & 11.5 & 72.1 \\
 & Percentile & 5.9 & 2.7 & 71.2 \\
 & Percentile + NBC & 6.8 & 3.3 & \textbf{71.7} \\
\midrule
\multirow{3}{*}{ViT-B} 
 & Full-precision & 346.3 & 84.3 & 84.5 \\
 & Percentile & 87.4 & 15.2 & 75.6 \\
 & Percentile + NBC & 101.6 & 18.4 & \textbf{83.5} \\
\midrule
\multirow{3}{*}{Swin-S} 
 & Full-precision & 198.4 & 60.0 & 83.2 \\
 & Percentile & 50.1 & 14.8 & 82.1 \\
 & Percentile + NBC & 58.0 & 17.7 & \textbf{82.9} \\
\bottomrule
\end{tabular}
\end{table}

\textbf{Effectiveness of BLT.}
As shown in Table~\ref{tab:nonlinear_transform}, we compare the performance of various nonlinear transformations (asinh, tanh, sigmoid, and our proposed BLT). Experimental results demonstrate that BLT achieves an excellent balance between accuracy and efficiency. In terms of accuracy, BLT consistently outperforms both tanh and sigmoid, while performing comparably to asinh.

However, asinh incurs noticeably higher inference latency on actual hardware than BLT, as summarized in Table~\ref{tab:asinh_latency}. We attribute this latency gap to the structural difference in their underlying operators. BLT is a piecewise composition of linear scaling and a $\log_2$ operation, both of which map directly to native GPU primitives, whereas asinh must be computed as $\log(x + \sqrt{x^2+1})$, requiring an additional square-root operation per element. On average across the three backbones, BLT introduces only $2.2$\,ms of additional latency over the W8A8 baseline, while asinh introduces $3.4$\,ms, approximately $1.5\times$ the deployment overhead of BLT. These latency measurements were conducted under the same experimental setup as detailed in Table~\ref{tab:latency_comparison}.

\begin{table}[t]
    \centering
    \begin{minipage}[t]{0.48\textwidth}
        \centering
        \caption{Comparison of Nonlinear Transformations. The accuracy is evaluated under 4-bit quantization, utilizing RepQ-ViT as the baseline method.}
        \label{tab:nonlinear_transform}
        \begin{tabular}{lcccc}
            \toprule
            Network & asinh & tanh & sigmoid & BLT \\
            \midrule
            DeiT-T & \textbf{61.9} & 61.8 & 60.6 & \textbf{61.9} \\
            ViT-B  & 77.5 & 74.5 & 74.5 & \textbf{77.6} \\
            Swin-S & 81.1 & 80.8 & 80.4 & \textbf{81.2} \\
            \bottomrule
        \end{tabular}
    \end{minipage}
    \hfill 
    \begin{minipage}[t]{0.48\textwidth}
        \centering
        \caption{Comparison of latency and Top-1 accuracy for asinh and BLT.}
        \label{tab:asinh_latency}
        \setlength{\tabcolsep}{4pt}
        \begin{tabular}{c c c c}
            \toprule
            Network & Transformation & Latency & Top-1 \\
            \midrule
            \multirow{2}{*}{DeiT-T} 
                & BLT & \textbf{3.3} & \textbf{71.7} \\
                & asinh & 3.8 & 71.6 \\
            \midrule
            \multirow{2}{*}{ViT-B} 
                & BLT & \textbf{18.4} & \textbf{83.5} \\
                & asinh & 21.1 & 83.0 \\
            \midrule
            \multirow{2}{*}{Swin-S}
                & BLT & \textbf{17.7} & 82.9 \\
                & asinh & 18.1 & \textbf{83.0} \\
            \bottomrule
        \end{tabular}
    \end{minipage}
\end{table}

\section{Conclusions and Limitations}

In this paper, we proposed a novel post-training quantization approach named Nonlinear Bipolar Compensation (NBC). We first demonstrated that existing compensation-based methods struggle to compensate for highly nonlinear quantization loss sufficiently and their performance is severely limited by outliers in activations. To address these issues, we proposed the NBC framework, which introduces nonlinearity compensation with minimal overhead. At its core, we designed Bipolar Logarithmic Transformation (BLT) to effectively suppress outliers and handle negative values. Extensive experimental results across vision, language, and multimodal tasks demonstrated the effectiveness, efficiency, robustness, and generality of our NBC method.

While NBC demonstrates strong effectiveness and efficiency, it currently has a few limitations. First, as a compensation-based method, NBC's performance is inherently dependent on existing PTQ frameworks. In future work, we aim to integrate NBC with next-generation PTQ algorithms to further push its performance ceiling. Second, the lack of a mature W4A4 inference framework for Vision Transformers makes end-to-end deployment under W4A4 currently infeasible. Integrating NBC with practical W4A4 deployment is left to future work.

\begin{ack}
This work was partly supported by the National Natural Science Foundation of China under Grant 62276123 and Fundamental and Interdisciplinary Disciplines Breakthrough Plan of the Ministry of Education of China (No. JYB2025XDXM118) , and the “111 Center” (No. B26023).

JW pointed out that outliers are a key factor limiting the performance of linear compensation methods and designed the NBC framework accordingly. PS developed these ideas into concrete algorithms and implementations, and completed the experimental validation. JW and PS wrote the paper.

We thank Ningyuan Tang and Jin Tong for helpful discussions.
\end{ack}

\section*{Supplementary Material}

\appendix

\section{Feature Loss based Local Search}
\label{app:fls}
Algorithm~\ref{alg:local_search_N} gives the full pseudocode of FLS for selecting the hyperparameter $N$. We employ a one-dimensional local search strategy. Inspired by the feature mimicking method~\cite{wang2021distilling}, we set the feature loss as the evaluation criterion. This loss is defined as the squared $L_2$-distance between the output features of the full-precision model and the quantized model just before the classification head, which is formalized as
\begin{equation}
\mathcal{L}_{mse} = \frac{1}{TD} \sum_{i=1}^{T} \sum_{j=1}^{D}\left\| \boldsymbol{F}_{i,j} - \boldsymbol{F}^q_{i,j} \right\|_2^2 \,,
\end{equation}
where $T$ denotes the total number of tokens, $D$ represents the feature dimensionality, $F$ is the original model's features, and $F^q$ represents the quantized model's features.

Starting from an empirical initial value $N_{\text{init}}$, we iteratively perform an exploratory search within the discrete neighborhood of $N$, specifically $\{N \pm s\}$, where $s$ is a preset step size. At each step, we evaluate the feature loss $\mathcal{L}_{\text{mse}}$ of the quantized model at that $N$ value. The search process continues until convergence to a local minimum $N_m$, which satisfies the following conditions:
\begin{align}
\mathcal{L}_{\text{mse}}(N_m) < \mathcal{L}_{\text{mse}}(N_m-s),  \\ \mathcal{L}_{\text{mse}}(N_m) < \mathcal{L}_{\text{mse}}(N_m+s).
\end{align}

Upon reaching $N_m$, the final chosen $N$ value is the one that yielded the minimum $\mathcal{L}_{\text{mse}}$ among all $N$ values explored during the entire search process. To prevent overfitting $(W, b)$ to the calibration data during the search, we adopt a hold-out protocol: the 512-image calibration set is split into 384 fit images and 128 hold-out images; for each candidate $N$, $(W, b)$ are fit on the former and $\mathcal{L}_{\text{mse}}$ is evaluated on the latter. Once the optimal $N_o$ is selected, $(W, b)$ are refit on all 512 images. Notably, our experimental results demonstrate that an appropriate hyperparameter $N$ can be found in a few steps with a well-chosen $N_{\text{init}}$ and $s$.

\begin{algorithm}[t]
\caption{Feature Loss based Local Search for $N$.}
\label{alg:local_search_N}
\begin{algorithmic}[1]
    \Statex \textbf{Input:} Initial value $N_{\text{init}}$; Minimum value $N_{\text{min}}$; Maximum value $N_{\text{max}}$; Step size $s$; Full-precision model $\mathcal{M}$; Quantized model $\mathcal{M}^q$.
    \Statex \textbf{Output:} Optimal hyperparameter $N_o$.
    \State Initialize an empty mapping $H$ and initialize an empty queue $Q$. Set $e$ to $False$.
    \State $\mathbf{Enqueue}(Q, N_{\text{init}})$.
    
    \If { $N_{\text{init}} + s \le N_{\text{max}}$ }
        \State $\mathbf{Enqueue}(Q, N_{\text{init}} + s)$.
    \EndIf
    \If { $N_{\text{init}} - s \ge N_{\text{min}}$ }
        \State $\mathbf{Enqueue}(Q, N_{\text{init}} - s)$.
    \EndIf

    \While {$Q$ is not empty}
    \State $N_{c} \gets \mathbf{Dequeue}(Q)$.
    \State Integrate NBC into $\mathcal{M}^q$ under the setting of $N_{c}$ and get $\mathcal{M}^c$.
    \State Calculate $\mathcal{L}_{\text{mse}}$ of $N_{c}$ using $\mathcal{M}^c$ and $\mathcal{M}$.
    \State $H[N_c] \leftarrow \mathcal{L}_{\text{mse}}$.

    \If { $N_c > N_{\text{init}} + s$}
    {
      \If { $N_c - s $ is local minimum}
        {
            Set $e$ to $True$.
        }
        \EndIf
    }
    \EndIf

    \If { $N_c < N_{\text{init}}$}
    {
      \If { $N_c + s $ is local minimum}
        {
          Set $e$ to $True$.
        }
        \EndIf
    }
    \EndIf

    \If  {not $e$}
    {
        \If { $N_c > N_{\text{init}}$ and $N_{c} + s \le N_{\text{max}}$}
        {
          $\mathbf{Enqueue}(Q, N_{c} + s)$.
        }
        \ElsIf{$N_c < N_{\text{init}}$ and $N_{c} - s \ge N_{\text{min}}$}
        {
          $\mathbf{Enqueue}(Q, N_{c} - s)$.
        }
        \EndIf
    }
    \EndIf
    \EndWhile
\State $N_o \gets \arg\min_{N} \left\{ H[N] \right\}$.
\end{algorithmic}
\end{algorithm}

\textbf{Effectiveness and efficiency of FLS.} We conducted experiments to verify the effectiveness and efficiency of FLS. In our experiments, we adopted a search setting with $N_{\text{init}} = 2$ and $s = 1$. The detailed results are shown in Table~\ref{tab:fls_N}. The results demonstrate that FLS can quickly find a suitable $N$ within a few steps. For example, it requires searching only 4 values of $N$ for ViT-B and 5 for Swin-S.

\begin{table}[bp]
    \centering
    \caption{Ablation on FLS under 4-bit quantization on ImageNet~\cite{deng2009imagenet} using the ViT-B~\cite{dosovitskiy2020image} and Swin-S~\cite{liu2021swin} backbones. The finally selected $N$ and its corresponding feature loss and top-1 accuracy are shown in bold.}
    \label{tab:fls_N}
    \begin{tabular}{l c c c}
        \toprule
        Network & $N$ & Feature Loss & Top-1 \\
        \midrule
        \multirow{4}{*}{ViT-B}
            & 1 & 6.4235 & 77.4 \\
            & \textbf{2} & \textbf{6.0595} & \textbf{77.6} \\
            & 3 & 6.3135 & 76.9 \\
            & 4 & 6.2044 & 77.0 \\
        \midrule
        \multirow{5}{*}{Swin-S}
            & 0 & 0.0579 & 80.9 \\
            & 1 & 0.0553 & 81.0 \\
            & 2 & 0.0537 & 81.1 \\
            & \textbf{3} & \textbf{0.0536} & \textbf{81.2} \\
            & 4 & 0.0562 & 81.1 \\
        \bottomrule
    \end{tabular}
\end{table}

\textbf{Robustness and convergence of FLS.} We further investigated the influence of the step size $s$ on both the search results for hyperparameter $N$ and the final accuracy. The results are shown in Table~\ref{tab:fls_s}. From the experimental results, we observe that as $s$ decreases, the final accuracy increases slightly. However, a smaller $s$ often requires a longer search time. We used $s=1$ in all our experiments. 

\begin{table}[tbp]
    \centering
    \caption{Ablation on $s$ in BLT under 4-bit quantization on ImageNet~\cite{deng2009imagenet} using the ViT-B~\cite{dosovitskiy2020image} and Swin-S~\cite{liu2021swin} backbones. `Chosen $N$' refers to the value of $N$ ultimately selected by FLS.}
    \label{tab:fls_s}
    \begin{tabular}{l c c c}
        \toprule
        Network & $s$ & Chosen $N$ & Top-1 \\
        \midrule
        \multirow{3}{*}{ViT-B}
            & 0.5 & 2.0 & 77.6 \\
            & 1.0 & 2.0 & 77.6 \\
            & 2.0 & 2.0 & 77.6 \\
        \midrule
        \multirow{3}{*}{Swin-S}
            & 0.5 & 3.0 & 81.2 \\
            & 1.0 & 3.0 & 81.2 \\
            & 2.0 & 4.0 & 81.1 \\
        \bottomrule
    \end{tabular}
\end{table}

In our experiments, we set $N_{\min} = -10$ and $N_{\max} = 10$. The initial value $N_{\text{init}}$ is set to $2$, a choice grounded in the empirical observation that this value corresponds to approximately 50\% of the activation distribution in the ViT-B model. This statistical initialization allows FLS to efficiently explore a locally convex neighborhood.

\section{Why Linear Compensation Fails on Outliers}
\label{app:ols_outlier}

We use the notation of Equation~\ref{eq:qwt} to formalize the failure mode discussed in Section~\ref{sec:intro}. Since the multidimensional OLS problem decouples row-wise, we analyze the scalar case: a single entry $w$ of $W$ regresses one coordinate of the quantization error $r = y - y^q$ on one coordinate of the quantized input $x^q$ (ignoring the bias does not affect the argument). The scalar OLS problem $\min_w \sum_i (r_i - w x_i^q)^2$ admits the closed-form solution
\begin{equation}
    w^* = \frac{\sum_i x_i^q r_i}{\sum_i (x_i^q)^2}.
\end{equation}
Consider an idealized calibration set with $K$ outliers ($|x_i^q| = M$) and $n - K$ normal samples ($|x_i^q| = m \ll M$), whose average quantization errors are $\bar{r}_o$ and $\bar{r}_n$ respectively. Substituting into the closed-form solution gives
\begin{equation}
    w^*
    \approx
    \frac{K M \bar{r}_o + (n-K) m \bar{r}_n}{K M^2 + (n-K) m^2}.
\end{equation}
This expression reveals two sources of bias. (i) Denominator domination. When $M \gg m$, the denominator is dominated by the $K M^2$ term, so $w^*$ is governed by the outlier group rather than the normal majority. (ii) Numerator amplification. If the outlier group also has a large average error $\bar{r}_o$, the numerator term $K M \bar{r}_o$ further pulls $w^*$ toward the outliers. Together, these two effects explain the observation in Section~\ref{sec:key_ideas}: outliers in both $x^q$ and $y - y^q$ pose severe challenges to linear compensation.

\section{Additional Ablation Study}
\textbf{Ablation on calibration set size.}
We conducted experiments on different calibration set sizes. Table~\ref{tab:calibration_size} reports the experimental results on the ViT-B~\cite{dosovitskiy2020image} and Swin-S~\cite{liu2021swin} backbones.

The results show that ViT-B is more sensitive to changes in the calibration set size, with its accuracy improving by 4.4\% when increasing the calibration set size from 32 images to 1024 images. In contrast, Swin-S exhibits minimal dependency on the calibration set size, showing only a 0.3\% accuracy difference between 32 and 1024 images, and reaching its peak performance at 512 images. These results demonstrate that selecting 512 images as the calibration set achieves a balance between efficiency and accuracy.

\begin{table}[t]
\centering
\caption{Ablation on calibration set size of NBC on ImageNet~\cite{deng2009imagenet} classification.} 
\label{tab:calibration_size} 
\begin{tabular}{ll c}
\toprule
Network & Calibration Size & Top-1 \\
\midrule
\multirow{4}{*}{ViT-B}
& 32 images & 73.4 \\
& 128 images & 76.4 \\
& 512 images & 77.6 \\
& 1024 images & 77.8 \\
\midrule
\multirow{4}{*}{Swin-S}
& 32 images & 80.9 \\
& 128 images & 81.1 \\
& 512 images & 81.2 \\
& 1024 images & 81.2 \\
\bottomrule
\end{tabular}
\end{table}

\textbf{Importance of transforming both quantized inputs and quantization errors.} We conducted an ablation experiment to compare three different transformation strategies: (1) BLT transformation is applied only to the quantized input $x^q$; (2) transformation is applied only to the quantization error term $y - y^q$; and (3) transformation is applied to both $x^q$ and $y - y^q$ simultaneously (i.e., the NBC transformation strategy).

Table~\ref{tab:abl_transformation} presents the experimental results for the 4-bit quantization setting on ViT-B~\cite{dosovitskiy2020image} and Swin-S~\cite{liu2021swin}. The experimental results show that simultaneously transforming both the quantized input and the quantization error yields the best accuracy, validating the effectiveness of the NBC transformation strategy.

\begin{table}
    \centering
    \caption{Ablation on the transformation settings under 4-bit quantization on ImageNet~\cite{deng2009imagenet} using the ViT-B~\cite{dosovitskiy2020image} and Swin-S~\cite{liu2021swin} backbones. `Transformation Setup' describes the three transformation settings used in the experiments: transforming only the quantized input; transforming only the quantization error; and transforming both the quantized input and the quantization error simultaneously. }
    \label{tab:abl_transformation}
    \begin{tabular}{l c c}
        \toprule
        Network & Transformation Setup & Top-1  \\
        \midrule
        \multirow{3}{*}{ViT-B}
            & Input \& Error & \textbf{77.6} \\
            & Only Input           & 76.9 \\
            & Only Error           & 77.5 \\
        \midrule
        \multirow{3}{*}{Swin-S}
            & Input \& Error & \textbf{81.2} \\
            & Only Input           & 81.0 \\
            & Only Error           & 80.7 \\
        \bottomrule
    \end{tabular}
\end{table}

\textbf{Compare NBC against QwT-v2.}
We compare NBC with the QwT-v2~\cite{tang2025qwt} method in Table~\ref{tab:cls_qwtv2}. The results show that NBC significantly outperforms QwT-v2 across all backbones. Specifically, NBC achieves a +2.0\% Top-1 accuracy improvement on both ViT-B and DeiT-T.

We acknowledge that NBC leads to a slightly larger model size compared to QwT-v2. However, this is due to the different research priorities: while QwT-v2 focuses on extreme parameter efficiency with minimal storage overhead, our NBC prioritizes maximizing accuracy recovery in challenging 4-bit scenarios. The substantial accuracy gains (e.g., +2.0\% on ViT-B) justify the modest increase in size, making NBC a more effective solution for accuracy-critical deployments.

\begin{table}[tbp]
    \centering
    \caption{Comparison with QwT-v2 on ImageNet~\cite{deng2009imagenet} classification under 4-bit quantization.}
    \label{tab:cls_qwtv2}
    \begin{tabular}{ll c c c}
        \toprule
        Network & Method & \#Bits & Size & Top-1 \\
        \midrule
        \multirow{4}{*}{DeiT-T} & Full-precision & 32/32 & 22.9 & 72.1 \\
        \cdashline{2-5}
        & RepQ-ViT~\cite{li2023repq} & 4/4 & 3.3 & 57.6 \\
        & RepQ-ViT + QwT-v2 & 4/4 & 3.4 & 59.9 \\
        & RepQ-ViT + NBC & 4/4 & 4.2 & \textbf{61.9} \\
        \midrule
        \multirow{4}{*}{ViT-B} & Full-precision & 32/32 & 346.3 & 84.5 \\
        \cdashline{2-5}
        & RepQ-ViT~\cite{li2023repq} & 4/4 & 44.9 & 69.1 \\
        & RepQ-ViT + QwT-v2 & 4/4 & 45.6 & 75.6 \\
        & RepQ-ViT + NBC & 4/4 & 59.1 & \textbf{77.6} \\
        \midrule
        \multirow{4}{*}{Swin-S} & Full-precision & 32/32 & 198.4 & 83.2 \\
        \cdashline{2-5}
        & RepQ-ViT~\cite{li2023repq} & 4/4 & 25.8 & 80.2 \\
        & RepQ-ViT + QwT-v2 & 4/4 & 26.4 & 80.3 \\
        & RepQ-ViT + NBC & 4/4 & 33.7 & \textbf{81.2} \\
        \bottomrule
    \end{tabular}
\end{table}

\textbf{Robustness to calibration-set sampling.} To assess the statistical reliability of the single-seed results reported in the main paper, we re-ran the 4-bit quantization pipeline with 5 different random seeds for the calibration-set sampling on three representative backbones. Table~\ref{tab:err_bar} reports the mean and standard deviation of top-1 accuracy on ImageNet for both RepQ-ViT and RepQ-ViT + NBC.

\begin{table}[t]
\centering
\caption{Top-1 accuracy (\%) on ImageNet under 4-bit quantization across 5 random seeds for the calibration-set sampling. Mean $\pm$ standard deviation is reported. ``Lift'' denotes the accuracy improvement achieved by NBC over the RepQ-ViT baseline.}
\label{tab:err_bar}
\begin{tabular}{lccc}
\toprule
Model & RepQ-ViT & RepQ-ViT + NBC & Lift \\
\midrule
DeiT-T & $57.84 \pm 0.19$ & $61.71 \pm 0.27$ & $+3.87$ \\
ViT-B  & $68.47 \pm 0.75$ & $77.19 \pm 0.44$ & $+8.72$ \\
Swin-S & $80.06 \pm 0.10$ & $81.16 \pm 0.06$ & $+1.10$ \\
\bottomrule
\end{tabular}
\end{table}

NBC delivers a consistent mean accuracy lift across all three backbones, and on each backbone the lift exceeds the seed-induced standard deviation, confirming that the gains reported in the main paper are not artifacts of a particular calibration-set sample.

\textbf{Calibration efficiency of NBC.} To support our claim that NBC completes calibration within a few minutes, we report the total calibration time, which includes the FLS search for the hyperparameter $N$ and the closed-form computation of the NBC parameters. Measurements were taken on a single NVIDIA RTX 3090 GPU under 4-bit quantization with a calibration set of 512 ImageNet~\cite{deng2009imagenet} images. As shown in Table~\ref{tab:calibration_time}, the entire calibration finishes in roughly one to three minutes across all six backbones. This confirms that NBC preserves the practical efficiency of calibration-only PTQ methods, whereas optimization-based methods typically require hours of blockwise reconstruction.

\begin{table}[t]
\centering
\caption{End-to-end calibration time under 4-bit quantization on ImageNet~\cite{deng2009imagenet} with a 512-image calibration set, measured on a single NVIDIA RTX 3090 GPU.}
\label{tab:calibration_time}
\begin{tabular}{lcccccc}
\toprule
Network & DeiT-T & DeiT-S & ViT-S & ViT-B & Swin-T & Swin-S \\
\midrule
Calibration Time (s) & 60.2 & 73.4 & 65.7 & 91.6 & 90.5 & 157.4 \\
\bottomrule
\end{tabular}
\end{table}

\section{Experiments on Object Detection \& Instance Segmentation}
\label{app:detection}

\textbf{Settings.} We evaluated our method on object detection and instance segmentation tasks using the COCO~\cite{lin2014microsoft} dataset. We employed Swin-S~\cite{liu2021swin} with Mask R-CNN~\cite{he2017mask} and Swin-S/B~\cite{liu2021swin} with Cascade Mask R-CNN~\cite{cai2018cascade} as detectors. We randomly selected 512 images from the training set to initialize the parameters of the NBC module and an additional 128 images to tune the hyperparameter $N$. We adopted RepQ-ViT~\cite{li2023repq} as the baseline PTQ algorithm.

\textbf{Main Results.} The experimental results of NBC on object detection and instance segmentation tasks are displayed in Table~\ref{tab:detection_main}. The results show that NBC improves $\text{AP}^{\text{box}}$ and $\text{AP}^{\text{mask}}$ in all cases, achieving an average gain of 0.3\% in $\text{AP}^{\text{box}}$ and 0.2\% in $\text{AP}^{\text{mask}}$. The baseline PTQ method reduces the model size to an average of 84\% of the full-precision model, and this reduction decreases to 81\% after integrating NBC.

\begin{table}[tbp]
    \centering
    \caption{Quantization results on COCO~\cite{lin2014microsoft} for object detection and instance segmentation models. We use box average precision ($\text{AP}^{\text{box}}$) and mask average precision ($\text{AP}^{\text{mask}}$) to evaluate the accuracy of object detection and instance segmentation, respectively.}
    \label{tab:detection_main}
    \setlength{\tabcolsep}{1pt}
    \begin{tabular}{ll c c c c}
        \toprule
        Network & Method & \#Bits & Size& $\mathbf{\text{AP}^{\text{box}}}$ & $\mathbf{\text{AP}^{\text{mask}}}$ \\
        \midrule
        \multirow{5}{*}{\shortstack{Swin-S \\ + Mask R-CNN}}
        & Full-precision & 32/32 & 276.5 & 48.5 & 43.3 \\
        \cdashline{2-6}
        & RepQ-ViT~\cite{li2023repq} & 4/4 & \phantom{2}36.1 & 42.8 & 40.0 \\
        & RepQ-ViT + NBC & 4/4 & \phantom{2}44.0 & \textbf{43.0} & \textbf{40.3} \\
        \cdashline{2-6}
        & RepQ-ViT~\cite{li2023repq} & 6/6 & \phantom{2}53.3 & 47.6 & 42.8 \\
        & RepQ-ViT + NBC & 6/6 & \phantom{2}61.2 & \textbf{47.8} & \textbf{43.0} \\
        \midrule
        \multirow{5}{*}{\shortstack{Swin-S \\ + Cascade \\Mask R-CNN}}
        & Full-precision & 32/32 & 427.8 & 51.9 & 45.0 \\
        \cdashline{2-6}
        & RepQ-ViT~\cite{li2023repq} & 4/4 & \phantom{2}56.9 & 49.1 & 43.0 \\
        & RepQ-ViT + NBC & 4/4 & \phantom{2}64.8 & \textbf{49.4} & \textbf{43.1} \\
        \cdashline{2-6}
        & RepQ-ViT~\cite{li2023repq} & 6/6 & \phantom{2}83.4 & 51.4 & 44.6 \\
        & RepQ-ViT + NBC & 6/6 & \phantom{2}91.3 & \textbf{51.7} & \textbf{44.8} \\
        \midrule
        \multirow{5}{*}{\shortstack{Swin-B \\ + Cascade \\ Mask R-CNN}}
        & Full-precision & 32/32 & 579.9 & 51.9 & 45.0 \\
        \cdashline{2-6}
        & RepQ-ViT~\cite{li2023repq} & 4/4 & \phantom{2}76.1 & 49.3 & 43.1 \\
        & RepQ-ViT + NBC & 4/4 & \phantom{2}90.1 & \textbf{49.6} & \textbf{43.4} \\
        \cdashline{2-6}
        & RepQ-ViT~\cite{li2023repq} & 6/6 & 112.1 & 51.5 & 44.8 \\
        & RepQ-ViT + NBC & 6/6 & 126.1 & \textbf{51.7} & \textbf{44.9} \\
        \bottomrule
    \end{tabular}
\end{table}

\section{Nonlinear Compensation using an MLP}
As described in Section~\ref{sec:intro}, we introduce a multilayer perceptron (MLP) to replace the linear compensation module for nonlinear compensation, which can be formalized as follows:
\begin{equation}
y^{\text{mlp}} = y^q + W_2 \cdot \sigma(W_1x^q + b_1) + b_2,
\label{mlp}
\end{equation}
where $W_1 \in \mathbb{R}^{h \times d_{in}}$ is the weight matrix and $b_1 \in \mathbb{R}^{h}$ is the bias vector for the first layer; $W_2 \in \mathbb{R}^{d_{out} \times h}$ is the weight matrix and $b_2 \in \mathbb{R}^{d_{out}}$ is the bias vector for the second layer; and $\sigma(\cdot)$ is the nonlinear activation function, which is specified as ReLU.

We use the gradient backpropagation algorithm to optimize the parameters $W_1$, $b_1$, $W_2$, and $b_2$ of the MLP. During the optimization process, only a small number of training samples, identical to those used by other compensation modules, are used as the training set to ensure the fairness of the comparison. The loss function used in training is as follows:
\begin{equation}
\mathcal{L}(y, y^{mlp}) = \frac{1}{M} \sum_{i=1}^{M} (y_i - y^{mlp}_i)^2,
\label{mlp_loss}
\end{equation}
where $M$ is the number of samples.

In the experiments for the MLP compensation module, we set the hidden layer dimension $h$ to $d_{in}/2$ and initialized the module parameters using the Xavier~\cite{glorot2010understanding} method. During the training process, we employed the dropout~\cite{srivastava2014dropout} technique to prevent overfitting. The module was trained for 4 epochs based on the loss function defined in Equation~\ref{mlp_loss}, employing a learning rate of 5e-4 and a dropout rate of 0.1. We set the batch size to 16 for DeiT~\cite{touvron2021training} and ViT~\cite{dosovitskiy2020image}, while for Swin~\cite{liu2021swin}, we set it to the total token count divided by 4096. Table~\ref{tab:mlp_results} shows the experimental results.

The significant accuracy improvement from the MLP compensation module illustrates the importance of nonlinear compensation. However, its requirement for training and tuning numerous hyperparameters limits its practical employment. Therefore, we propose the NBC method, which \emph{introduces nonlinearity while remaining simple and efficient}.

\begin{table}[tbp]
\centering
\caption{Quantization results on ImageNet~\cite{deng2009imagenet} classification. }
\label{tab:mlp_results}
\begin{tabular}{l c l c c c c}
\toprule
\multirow{2}{*}{Network} & \multirow{2}{*}{Full Prec.} & \multirow{2}{*}{Method} & \multicolumn{2}{c}{W4/A4} & \multicolumn{2}{c}{W6/A6} \\
\cmidrule(lr){4-5} \cmidrule(lr){6-7}
& & & Size & Top-1 & Size & Top-1 \\
\midrule
\multirow{2}{*}{DeiT-T} & \multirow{2}{*}{72.1} & RepQ-ViT~\cite{li2023repq} & 3.3 & 57.6 & 4.6 & 70.9 \\
& & RepQ-ViT + MLP & 4.2 & \textbf{62.3} & 5.5 & \textbf{71.2} \\
\midrule
\multirow{2}{*}{ViT-B} & \multirow{2}{*}{84.5} & RepQ-ViT~\cite{li2023repq} & 44.9 & 69.1 & 66.2 & 83.8 \\
& & RepQ-ViT + MLP & 59.1 & \textbf{76.8} & 80.4 & \textbf{83.9} \\
\midrule
\multirow{2}{*}{Swin-S} & \multirow{2}{*}{83.2} & RepQ-ViT~\cite{li2023repq} & 25.8 & 80.2 & 38.0 & 82.8 \\
& & RepQ-ViT + MLP & 19.0 & \textbf{81.3} & 26.1 & \textbf{83.0} \\
\bottomrule
\end{tabular}
\end{table}

\section{Implementation Details}
In this section, we describe the implementation details of our experiments. 

We implemented all experiments using PyTorch. All experiments were conducted using NVIDIA RTX 3090 GPUs. The image generation experiment was run on 8 GPUs. All other experiments, including image classification, object detection, instance segmentation, multimodal recognition, and large language models, were conducted on a single GPU.

\textbf{Image Classification.} We used RepQ-ViT~\cite{li2023repq} and AdaLog~\cite{wu2024adalog} as baseline methods. Following~\cite{li2023repq}, we randomly sampled 32 images from ImageNet~\cite{deng2009imagenet} for baseline method calibration. 

\textbf{Object Detection \& Instance Segmentation.}
Following~\cite{li2023repq}, we randomly sampled a single image from COCO~\cite{lin2014microsoft} for baseline method calibration. 

\textbf{Image Generation.} We randomly selected 512 images from the training set for NBC parameter initialization and FLS hyperparameter search. The other experimental settings are consistent with prior research~\cite{chen2025q}.

\textbf{Multimodal Recognition.} We randomly sampled 512 image-text pairs from the training data for NBC parameter initialization and FLS hyperparameter search.

\textbf{Large Language Models.} Following DuQuant~\cite{lin2024duquant}, the calibration set consists of 128 sequences of length 2048 from WikiText-2, used for both NBC parameter initialization and FLS hyperparameter search. Inspired by DWR~\cite{yu2025treasures}, we use perplexity on the calibration set, instead of feature loss, as the FLS criterion for our LLM experiments.

\section{More Experimental Results}
In this section, we provide more detailed experimental results. Table~\ref{tab:cls_deit}, Table~\ref{tab:cls_vit}, and Table~\ref{tab:cls_swin} report the image classification results on ImageNet across various backbones, including ViT~\cite{dosovitskiy2020image}, DeiT~\cite{touvron2021training}, and Swin~\cite{liu2021swin}.

The full results for the large language models are also shown in Table~\ref{tab:qa}. While the main text only reports the average accuracy across the five zero-shot commonsense QA datasets, here we present the individual results for each dataset.

\begin{table}[H]
\centering
\caption{Full results on ImageNet~\cite{deng2009imagenet} using DeiT~\cite{touvron2021training} backbone. }
\label{tab:cls_deit}
\begin{tabular}{l c l c c c c}
\toprule
\multirow{2}{*}{Network} & \multirow{2}{*}{Full Prec.} & \multirow{2}{*}{Method} & \multicolumn{2}{c}{W4/A4} & \multicolumn{2}{c}{W6/A6} \\
\cmidrule(lr){4-5} \cmidrule(lr){6-7}
& & & Size & Top-1 & Size & Top-1 \\
\midrule
\multirow{3}{*}{DeiT-T} & \multirow{3}{*}{72.1} & RepQ-ViT~\cite{li2023repq} & 3.3 & 57.6 & 4.6 & 70.9 \\
& & RepQ-ViT + QwT & 4.2 & 61.4 & 5.5 & 71.2 \\
& & RepQ-ViT + NBC & 4.2 & \textbf{61.9} & 5.5 & \textbf{71.2} \\
\midrule
\multirow{3}{*}{DeiT-S} & \multirow{3}{*}{79.9} & RepQ-ViT~\cite{li2023repq} & 11.9 & 69.0 & 17.2 & 78.9 \\
& & RepQ-ViT + QwT & 15.4 & 71.5 & 20.7 & \textbf{79.1} \\
& & RepQ-ViT + NBC & 15.4 & \textbf{72.5} & 20.7 & 78.9 \\
\bottomrule
\end{tabular}
\end{table}

\begin{table}[H]
\centering
\caption{Full results on ImageNet~\cite{deng2009imagenet} using ViT~\cite{dosovitskiy2020image} backbone. }
\label{tab:cls_vit}
\begin{tabular}{l c l c c c c}
\toprule
\multirow{2}{*}{Network} & \multirow{2}{*}{Full Prec.} & \multirow{2}{*}{Method} & \multicolumn{2}{c}{W4/A4} & \multicolumn{2}{c}{W6/A6} \\
\cmidrule(lr){4-5} \cmidrule(lr){6-7}
& & & Size & Top-1 & Size & Top-1 \\
\midrule
\multirow{3}{*}{ViT-S} & \multirow{3}{*}{81.4} & RepQ-ViT~\cite{li2023repq} & 11.9 & 65.4 & 17.2 & 80.5 \\
& & RepQ-ViT + QwT & 15.4 & 70.8 & 20.7 & \textbf{80.7} \\
& & RepQ-ViT + NBC & 15.4 & \textbf{71.1} & 20.7 & \textbf{80.7} \\
\midrule
\multirow{3}{*}{ViT-B} & \multirow{3}{*}{84.5} & RepQ-ViT~\cite{li2023repq} & 44.9 & 69.1 & 66.2 & 83.8 \\
& & RepQ-ViT + QwT & 59.1 & 76.6 & 80.4 & \textbf{83.9} \\
& & RepQ-ViT + NBC & 59.1 & \textbf{77.6} & 80.4 & \textbf{83.9} \\
\bottomrule
\end{tabular}
\end{table}

\begin{table}[H]
\centering
\caption{Full results on ImageNet~\cite{deng2009imagenet} using Swin~\cite{liu2021swin} backbone. }
\label{tab:cls_swin}
\begin{tabular}{l c l c c c c}
\toprule
\multirow{2}{*}{Network} & \multirow{2}{*}{Full Prec.} & \multirow{2}{*}{Method} & \multicolumn{2}{c}{W4/A4} & \multicolumn{2}{c}{W6/A6} \\
\cmidrule(lr){4-5} \cmidrule(lr){6-7}
& & & Size & Top-1 & Size & Top-1 \\
\midrule
\multirow{3}{*}{Swin-T} & \multirow{3}{*}{81.4} & RepQ-ViT~\cite{li2023repq} & 14.8 & 73.6 & 21.7 & 80.6 \\
& & RepQ-ViT + QwT & 19.2 & 75.4 & 26.0 & \textbf{80.7} \\
& & RepQ-ViT + NBC & 19.2 & \textbf{76.7} & 26.0 & \textbf{80.7} \\
\midrule
\multirow{3}{*}{Swin-S} & \multirow{3}{*}{83.2} & RepQ-ViT~\cite{li2023repq} & 25.8 & 80.2 & 38.0 & 82.8 \\
& & RepQ-ViT + QwT & 33.7 & 80.4 & 45.9 & \textbf{82.9} \\
& & RepQ-ViT + NBC & 33.7 & \textbf{81.2} & 45.9 & \textbf{82.9} \\
\bottomrule
\end{tabular}
\end{table}

\begin{table}[H]
\centering
\caption{Detailed results on common sense QA datasets.}
\label{tab:qa}
\setlength{\tabcolsep}{4pt}
\resizebox{\textwidth}{!}{
\begin{tabular}{l c c c c c c c c}
\toprule
Method & \#Bits & PIQA & ARC-E & ARC-C & BoolQ & HellaSwag & WinoGrande & QA. Avg \\
\midrule
Full-precision & 16/16 & 76.88 & 53.54 & 40.53 & 71.13 & 72.96 & 67.25 & 63.72 \\
\midrule
DuQuant & 4/4 & 75.24 & 51.89 & 36.77 & 67.86 & 69.54 & 62.12 & 60.57 \\
DuQuant + NBC & 4/4 & 75.57 & 52.06 & 38.82 & 70.76 & 69.24 & 62.35 &\textbf{ 61.47} \\
\bottomrule
\end{tabular}
}
\end{table}

\bibliographystyle{unsrtnat}
\bibliography{refs}

\end{document}